\documentclass{article}
\usepackage{spconf,amsmath,amsfonts,graphicx,multirow,xcolor,colortbl,hyperref, tablefootnote, multirow}

\title{Achieving RGB-D Level Segmentation Performance From a Single ToF Camera}
\name{\small{Pranav Sharma$^{3}$, Jigyasa Singh Katrolia$^{1}$, Jason Rambach$^1$, Bruno Mirbach$^1$, Didier Stricker$^{1,2}$, Juergen Seiler$^{3}$}}

\address{\small{$^1$German Research Center for Artificial Intelligence, Germany, $^2$RPTU Kaiserslautern, Germany, $^3$FAU Erlangen, Germany}}

\begin{document}

\maketitle

\begin{abstract}
   Depth is a very important modality in computer vision, typically used as complementary information to RGB, provided by RGB-D cameras. In this work, we show that it is possible to obtain the same level of accuracy as RGB-D cameras on a semantic segmentation task using infrared (IR) and depth images from a single Time-of-Flight (ToF) camera. In order to fuse the IR and depth modalities of the ToF camera, we introduce a method utilizing depth-specific convolutions in a multi-task learning framework. In our evaluation on an in-car segmentation dataset, we demonstrate the competitiveness of our method against the more costly RGB-D approaches.
\end{abstract}
\section{Introduction}
\label{sec:intro}

The research field of semantic segmentation is dominated by RGB images. Only recently it shifted in the direction of RGB-D semantic segmentation \cite{Hazirbas2016FuseNetID,depthawareCNN, cao2021shapeconv,Cheng2017deconvnet}. However, RGB images may not always be available due to practical, logistical and financial reasons. RGB-D cameras incur higher cost and more effort to calibrate the two cameras. Their larger package size often limits their place in real-world applications. Indeed, Time-of-Flight (ToF) depth cameras are often deployed without an accompanying RGB camera for applications like gesture control, in-car monitoring, industry automation and building management. Infrared images (IR) on the other hand are a by-product of ToF depth cameras (no additional sensor needed), but have not been explored sufficiently, specifically in combination with depth data \cite{agresti2017iccv,su2016materialclassification}.

\begin{figure}
    \centering
    \includegraphics[width=0.40\textwidth]{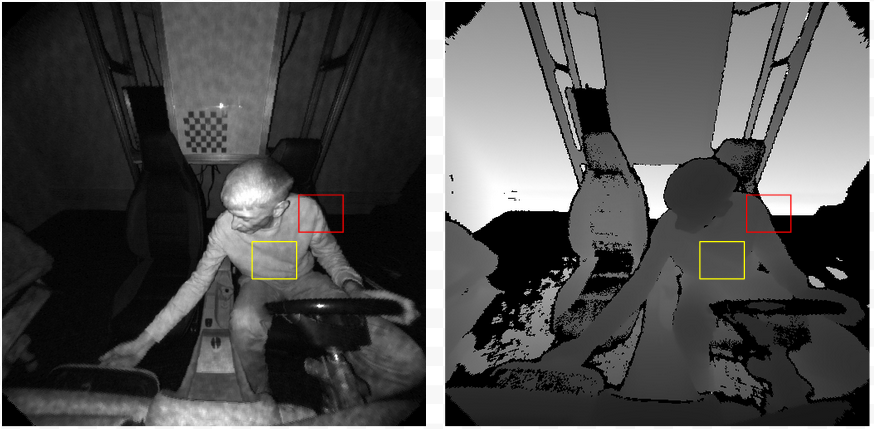}
    \caption{Patch similarities of IR and Depth modalities of a ToF Camera. }
    \label{fig:i-d_compare}
\end{figure}

Infrared images from ToF cameras provide the magnitude of the modulated light reflected from the scene and contain shape and semantic features in a different spectral range \cite{DBLP:phd/dnb/Hahne12}. Due to the similarities between RGB and IR images, it is natural to attempt to adapt existing RGB-D fusion approaches to combine IR and depth images. However, in most RGB-D methods depth information is only an accessory to the color information and is consumed by the same type of neural network layers despite their differences. Some recent works proposed depth-specific operations like depth-aware \cite{depthawareCNN} and shape-aware \cite{cao2021shapeconv} convolutions. We observe that both IR and depth outputs from a ToF camera are related in many ways and therefore these depth-specific operations can be applied to IR images as well. For example, the intensity of light reflected from an object decreases as distance to the object increases. Object surfaces closer to the camera will reflect more light implying that pixel intensities in an infrared image varies with the shape of that object. We can see this by comparing the same colored patches in Figure \ref{fig:i-d_compare} and note how both image patches have same relative changes in pixel values. We use this observation to leverage the shape-aware convolution operation for both IR and depth images to learn more meaningful features from both modalities. 

We aim to use the available modalities from a single ToF camera to achieve semantic segmentation performance comparable to RGB-D methods using an architecture that is tailored to IR-Depth (IR-D) input. We take inspiration from \cite{cao2021shapeconv,depthawareCNN} and design a depth-aware shape convolution operation that consumes IR-D input in a multi-task learning (MTL) architecture with depth filling as an auxiliary task. Our proposed method surpasses the baseline RGB-D based methods using only a ToF camera.

\section{Related Work}
\textbf{RGB-D Semantic Segmentation}:
 A wide range of approaches have been proposed for incorporating the depth information in RGB-D semantic segmentation and can be broadly classified into three categories as outlined by \cite{multi-modalsurvey}: (1) Using the depth channel as an additional input and performing fusion at different levels (early \cite{Song2017Fusion}, feature-level \cite{Hazirbas2016FuseNetID} or late \cite{Cheng2017deconvnet} fusion); (2) Depth as supervision signal for auxiliary tasks such as depth estimation or completion \cite{Jiao_2019_CVPR} and (3) Depth-specific operations. Depth-Aware CNN \cite{depthawareCNN} showed that object boundaries correlate with depth gradients and created \textit{depth-aware convolution} and \textit{depth-aware pooling} functions. ShapeConv \cite{cao2021shapeconv} used a \textit{shape convolution} kernel to ensure convolution kernels give consistent responses to object classes at different locations in the scene.  \cite{Chen20193DNC} used depth information to adjust the neighbourhood size of a 3D convolution filter.

\textbf{Depth and IR for Semantic Segmentation.}
Depth-only methods for semantic segmentation are typically applied to solve very specific tasks, like object manipulation via mechanical arm \cite{stacked-object-seg}, hand segmentation or hand and object segmentation \cite{rezaei2021weakly,hand-and-object-seg}. Even fewer methods have explored the combination of IR and depth. \cite{su2016materialclassification} used IR to classify materials since the infrared response depends on the material of the object. \cite{agresti2017iccv} used IR image as an additional input channel for improving the depth predicted from a setup containing both stereo and ToF cameras 

\textbf{Multi-Task Learning (MTL).}
The survey by \cite{MTLSurvey} describes many examples where simultaneous learning of two or more related tasks can boost the performance on either task. RGB-D segmentation is enhanced using auxiliary tasks like depth and surface normal prediction \cite{MTL1}. MTL methods typically employ a single encoder to learn features from the available input modalities and two separate task-specific decoders to perform prediction \cite{MTL1}. 

\section{Background and Notation}

\subsection{ShapeConv: Shape-aware Convolutional Layer}\label{sec:shapeconv}

In order to design convolutions that are invariant to different depth values (\textit{base}) when the underlying relative difference in depth in a local patch (\textit{shape}) remains same, ShapeConv \cite{cao2021shapeconv} suggested decomposing an image patch $\mathbb{P} \in R^{K_h\text{x}K_w\text{x}C_{in}}$ into a \textit{shape} component $\mathbb{P}_s$  and a \textit{base} component $\mathbb{P}_b$. These two components are operated on separately by a corresponding shape $\mathbb{W}_S$ and base kernel $\mathbb{W}_B$ before being passed to the standard convolutional kernel $\mathbb{K}$. Instead of decomposing the image patch, the convolutional kernel itself can be decomposed to the respective components as shown in equation \eqref{eq:shapeconv2}. Here $m(\mathbb{K})$ refers to mean of the kernel.
\begin{align}
\label{eq:shapeconv2}
\begin{split}
\mathbb{K}_B &= m(\mathbb{K})
\\
\mathbb{K}_S &= \mathbb{K} - m(\mathbb{K})
\end{split}
\end{align}

Shape convolution is then written with $ShapeConv$ as: 
\begin{align}
\label{eq:shapeconv3}
\begin{split}
\mathbb{F} &= ShapeConv(\mathbb{K}, \mathbb{W}_B, \mathbb{W}_S, \mathbb{P}_i)
\\
&=Conv(\mathbb{W}_B \diamond m(\mathbb{K}) + \mathbb{W}_S * (\mathbb{K} - m(\mathbb{K})), \mathbb{P})
\\
&=Conv(\mathbb{W}_B \diamond \mathbb{K}_B + \mathbb{W}_S * \mathbb{K}_S, \mathbb{P})
\\
&=Conv(\mathbf{K_B} + \mathbf{K_S}, \mathbb{P})
\\
&=Conv(\mathbf{K_{BS}}, \mathbb{P})
\end{split}
\end{align}

Here, $\diamond$ and $*$ represents the base and shape product respectively. $\mathbb{W}_B$ and $\mathbb{W}_S$ are learnable weights corresponding to \textit{base} and \textit{shape} components respectively.


\subsection{Depth-Aware CNN and Depth Similarity}
\label{subsection_dcnn}
In depth-aware convolution DCNN \cite{depthawareCNN}, pixels with similar depth values to the centre pixel are weighted more than other pixels. This property is named depth similarity. The depth similarity function $\text{F}_D(p_i,p_j)$ calculates the difference of depth values between two pixels $p_i$ and $p_j$ respectively. 
\begin{equation}
    \text{F}_D(p_i,p_j) = \text{exp}(-\alpha |D(p_i) - D(p_j)|)
    \label{eq:eqdcnn3}
\end{equation}

Depth-aware convolution is written with the depth similarity function $\text{F}_D$ as:\\
\begin{equation}
    y(p_0) = \sum_{p_n \in R}w(p_n)\text{F}_D(p_0, p_0 + p_n)x(p_0 + p_n)
    \label{eq:eqdcnn2}
\end{equation}
In Equation \eqref{eq:eqdcnn2}, the depth similarity term ($\text{F}_D$) is introduced with the convolution operation. The convolved features are weighted by $\text{F}_D$ . The parameter $\alpha$ weighs the influence of the depth similarity function $\text{F}_D$ on the convolution operation.

\section{Method}
We propose a depth-aware shape convolution operation and use it in a multi-task learning network. Our primary task is semantic segmentation using concatenated infrared and depth images from a ToF camera and the auxiliary task is depth completion for missing pixels in raw depth images. 

\subsection{Depth-Aware Shape Convolution} 
\label{Depth-aware Shape Convolution}
We design a depth-aware shape convolution, where the shape kernel in ShapeConv is supplemented with the depth similarity measure $\text{F}_D$ as computed in equation \eqref{eq:eqdcnn3}. Formally, this integration can be written in two steps. First the kernel is decomposed into \textit{shape} and \textit{base} kernel as shown in equation \eqref{eq:shapeconv3}. After the calculation of the weights $\mathbf{K_{BS}}$, the term $w(p_n)$ in equation \eqref{eq:eqdcnn2} is replaced with $\mathbf{K_{BS}}$ calculated from shape kernel. In this way, kernel weights calculated using shape kernel are integrated with depth similarity of DCNN. Equation \eqref{eq:eqdcnn3} can thus be rewritten for Depth-aware ShapeConv as:\\
\begin{equation}
     y(p_0) = \sum_{p_n \in R}\mathbf{K_{BS}}_{n}\text{F}_D(p_0, p_0 + p_n)x(p_0 + p_n)
     \label{eqIR_D1}
\end{equation}

\subsection{Infrared and Depth-Aware Multi-Task Network}

\begin{figure}
    \centering
    \includegraphics[width=0.43\textwidth]{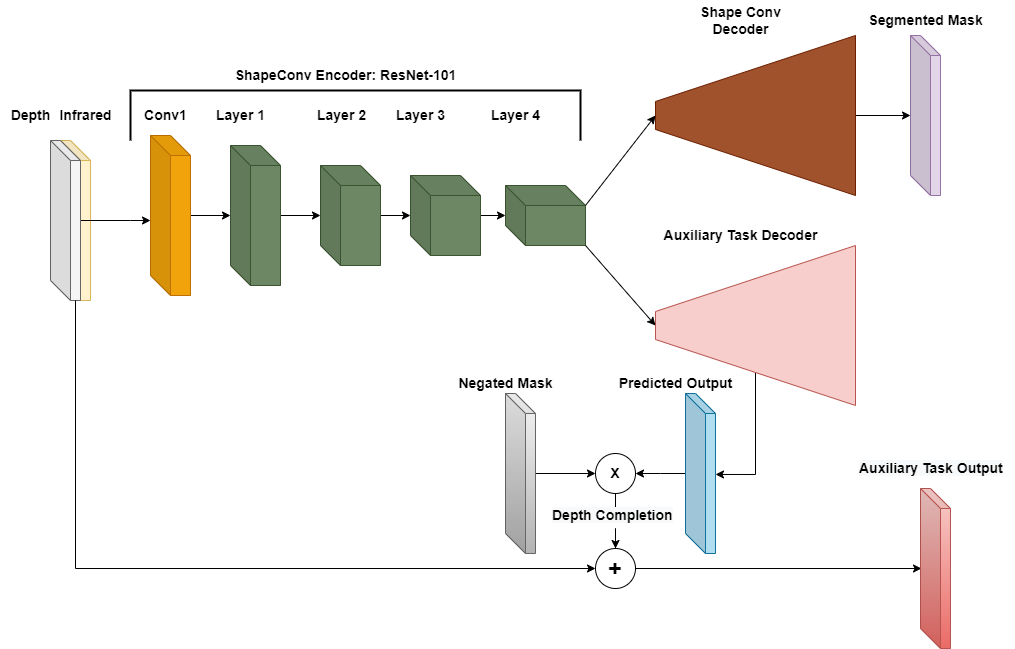}
    \caption{MTL architecture for segmentation task and dense depth prediction using depth-aware shape convolutions.}
    \label{fig:arch_mtl1}
\end{figure}

We realize a hard parameter sharing-based multi-task network with semantic segmentation as the main task and depth-filling as the auxiliary task (Figure \ref{fig:arch_mtl1}). We use ResNet-101 as the backbone feature extractor to encode features from concatenated infrared and depth (IR-D) images. The convolution layers in the ResNet encoder are replaced with depth-aware shape convolutions presented in section \ref{Depth-aware Shape Convolution}. The extracted features are passed to two task-specific decoders that generate final segmentation masks and depth values for missing pixels. For training the depth filling branch, the ground truth is prepared as described in section \ref{subsection_dataset}. We follow the training strategy from \cite{MTLPostprocessing} and use predicted depth values only for missing pixels to calculate the error between ground truth and predicted depth. The dense depth map predicted by the network is then multiplied with the missing pixels mask (1 for missing pixels, 0 otherwise) to keep predicted depth values only for the pixels that are missing in raw image. The remaining values are then replaced by the corresponding depth values in input image. 

\section{Experiments}
\subsection{Dataset} 
\label{subsection_dataset}
We evaluate our approach on the in-car cabin dataset TICaM \cite{katrolia2021ticam} that provides RGB, depth and infrared images recorded with a single ToF camera and corresponding segmentation masks. Surprisingly we could not find any other dataset that provided segmentation masks for all three image modalities: RGB, depth and infrared. We used the real image-set with the suggested split of 4666 training images and 2012 test images for our experiments. Following \cite{katrolia2021ticam} we combine different object classes into a single 'object' class to have 6 object classes in total. The RGB images have different resolution and FoV to depth and infrared images. To align the images, the RGB images are first mapped to the pinhole model of the depth images. Subsequently all the images are centre-cropped to size $230\times418$. Normalization of the infrared image is implemented by first removing the outliers by calculating 99\textsuperscript{th} percentile of the image and then scaling the image to the range of 0-255. To further enrich the information, histogram equalization and gamma filtering are applied to the normalized infrared image.


\begin{table*}[h]
\caption{Segmentation baselines for RGB-D and IR-D data. The input is always 4-channel with 3-channel RGB or infrared.}\label{traintest} \centering
\begin{tabular}{p{1.5cm}|p{3.5cm}|p{1.5cm}|p{1.5cm}|p{1.5cm}|p{1.5cm}}
\hline
\textbf{Input} & \textbf{Baselines} & \textbf{Pixel Acc.} & \textbf{Class Acc.} & \textbf{Mean IoU} & \textbf{f.w.IoU} \\ 
\hline
\multirow{3}{*}{RGB-D} & ShapeConv \cite{cao2021shapeconv} & \textbf{97.86} & \textbf{81.25} & \textbf{77.39} & \textbf{95.92} \\ 
  & Depth-Aware CNN \cite{depthawareCNN} & 94.63 & 66.88 & 54.14 & 90.78 \\ 
  & FuseNet \cite{Hazirbas2016FuseNetID} & 95.35 & 56.89 & 42.46 & 92.43 \\ 
\hline
\multirow{3}{*}{IR-D} & ShapeConv \cite{cao2021shapeconv} & \textbf{97.75} & \textbf{81.31} & \textbf{74.61} & \textbf{
95.76} \\
& Depth-Aware CNN \cite{depthawareCNN} & 93.52  & 60.52 & 50.03 & 88.45 \\ 
& FuseNet \cite{Hazirbas2016FuseNetID} & 93.18  & 55.73 & 39.61 & 88.84 \\ 
\hline
\end{tabular}
\label{baselineresults}
\end{table*}

\textbf{Ground-truth preparation.} To train auxiliary task of dense depth prediction, filled depth images are required as ground truth. As TICaM dataset only provides depth images with holes, filled versions of these depth maps are artificially created in this work using "Colorization using Optimization" scheme \cite{Colorizationscheme}. The depth images are filled by enforcing similar depth values to the neighbouring pixels with similar intensities. The information on pixel intensities is provided by infrared images. During training, the error is calculated for the missing pixels following the training strategy of \cite{MTLPostprocessing}.
\subsection{Baselines}
We choose three existing methods for RGB-D segmentation and train them on both RGB-D and IR-D images to establish our baselines and better evaluate the difference between RGB-D and IR-D inputs. We choose FuseNet \cite{Hazirbas2016FuseNetID} since it is an established and well-tested network on many benchmark datasets for RGB-D segmentation, however it has not been tested yet on TICaM dataset. ShapeConv and Depth-aware CNN on the other hand are more recent methods that use novel convolution operations unlike FuseNet. FuseNet and Depth-aware CNN have VGG-16 backbone, while ShapeConv uses ResNet-101. All networks use SGD optimizer with a momentum of 0.9 and weight decay of $5\times10^{-4}$ to update the weights. By default, Deeplab v3+ and ShapeConv use pre-trained weights while FuseNet and Depth-aware CNN are initialized using kaiming initialization \cite{kaiming}. For training with IR-D images we replicate infrared images to form 3-channel images and concatenate them with single channel depth images. We report pixel accuracy, class accuracy, mean Intersection-over-Union (IoU) and frequency weighted IoU (f.w.IoU) in Table \ref{baselineresults} for all the baselines.

\begin{figure}[htb]
   	\centering
	\setlength{\tabcolsep}{2pt}
 	\begin{tabular}{cccccccc}

 		\rotatebox{90}{\tiny{Depth}} &\includegraphics[scale=0.07875]{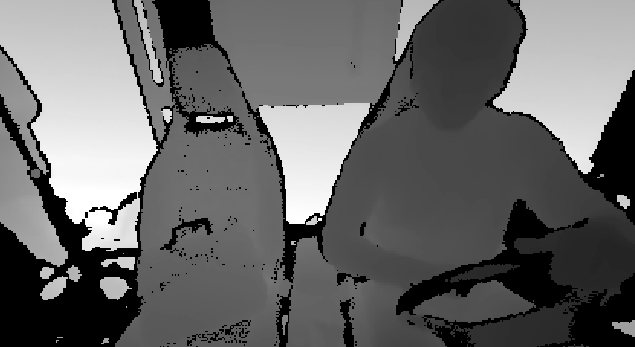}&
 		\includegraphics[scale=0.07875]{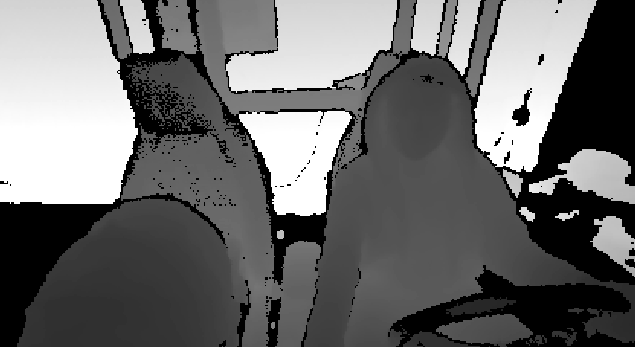}& 
 		\includegraphics[scale=0.07875]{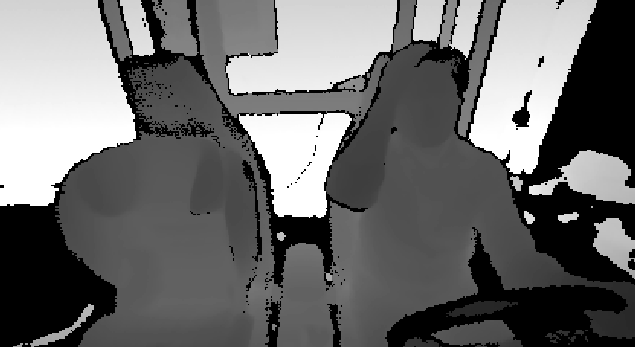}&
 		\includegraphics[scale=0.07875]{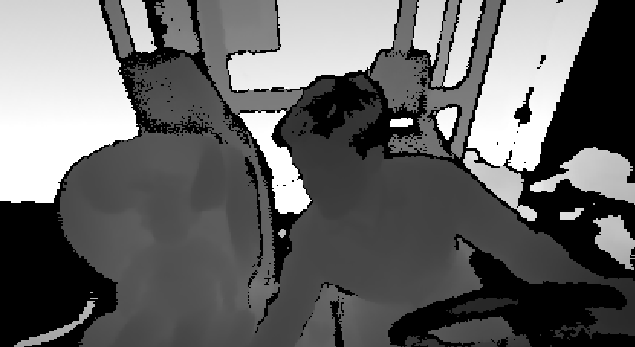}
 		\\
 		\rotatebox{90}{\tiny{Infrared}} &\includegraphics[scale=0.1665]{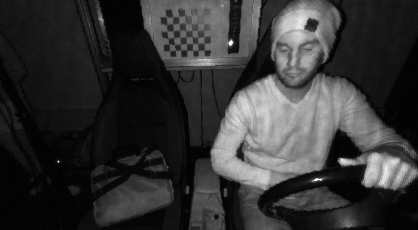}& 
 		\includegraphics[scale=0.1665]{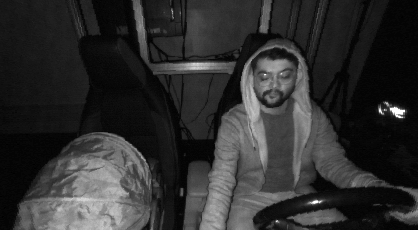}& 
 		\includegraphics[scale=0.1665]{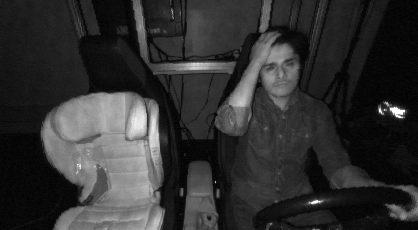}& 
 		\includegraphics[scale=0.1665]{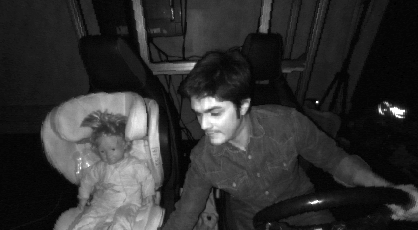}
 		\\
 		\rotatebox{90}{\tiny{Ground}} \rotatebox{90}{\tiny{Truth}} &\includegraphics[scale=0.1665]{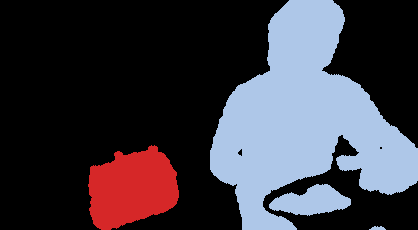}& 
 		\includegraphics[scale=0.1665]{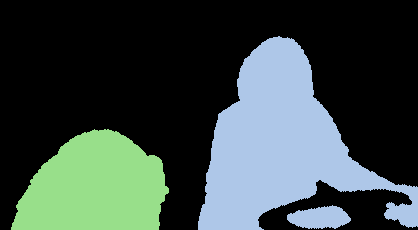}&
 		\includegraphics[scale=0.1665]{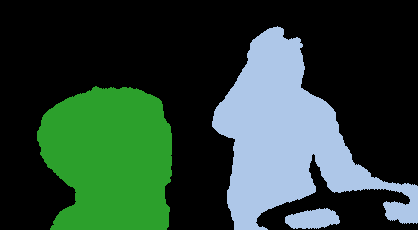}& 
 		\includegraphics[scale=0.1665]{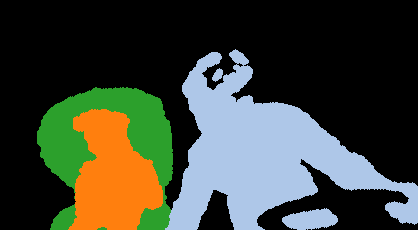}
       \\
 		\rotatebox{90}{\tiny{ShapeConv}} \rotatebox{90}{\tiny{RGB-D}} & \includegraphics[scale=0.1665]{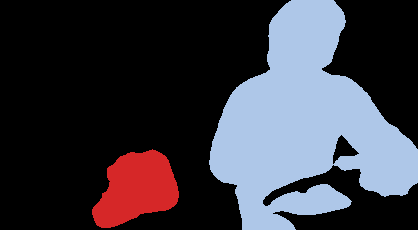}&
 		\includegraphics[scale=0.1665]{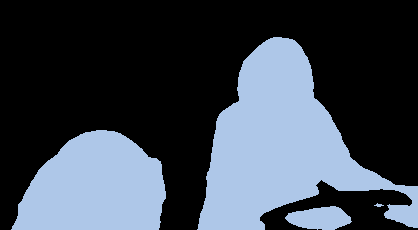}& 
 		\includegraphics[scale=0.1665]{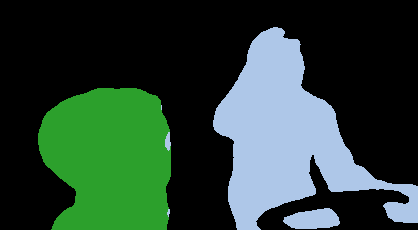}& 
 		\includegraphics[scale=0.1665]{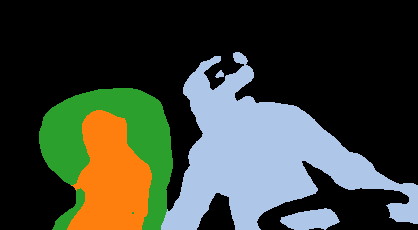}
 		\\
 		\rotatebox{90}{\tiny{ShapeConv}} \rotatebox{90}{\tiny{3ch IR-D}} & \includegraphics[scale=0.07875]{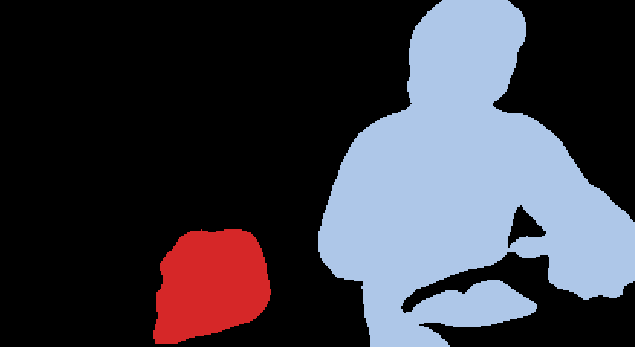}& 
 		\includegraphics[scale=0.07875]{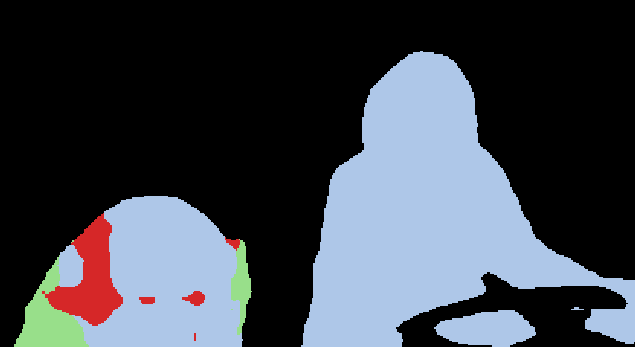}& 
 		\includegraphics[scale=0.07875]{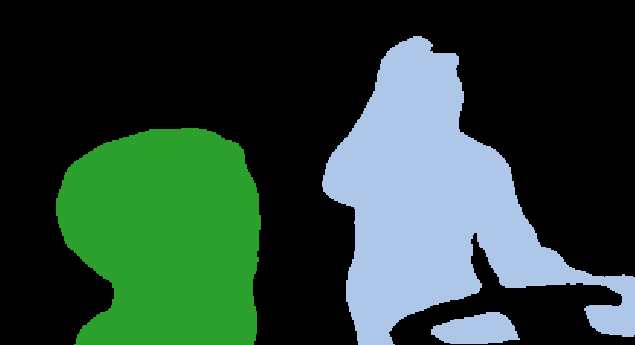}&
 		\includegraphics[scale=0.07875]{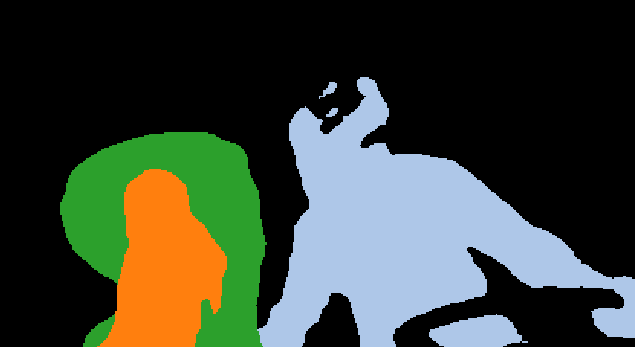}
 		\\

 		\rotatebox{90}{\tiny{ShapeConv}} \rotatebox{90}{\tiny{1ch IR-D}} & \includegraphics[scale=0.07875]{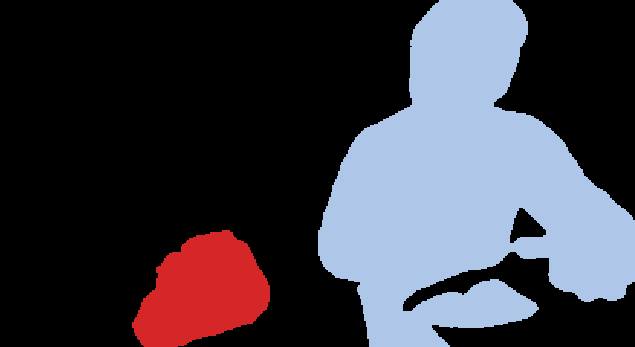}& 
 		\includegraphics[scale=0.07875]{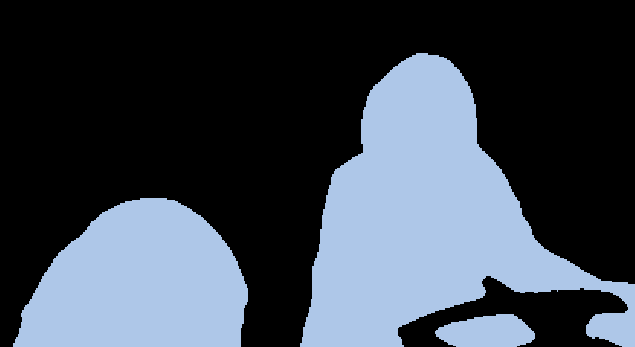}& 
 		\includegraphics[scale=0.07875]{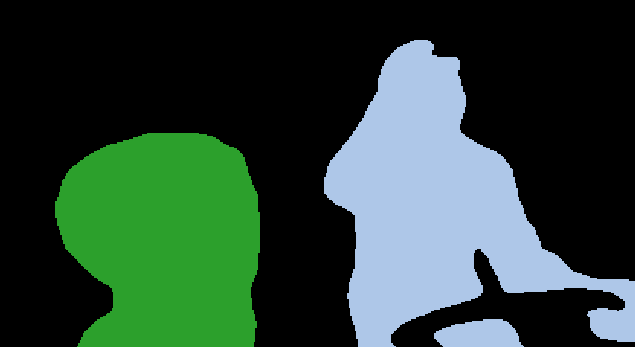}& 
 		\includegraphics[scale=0.07875]{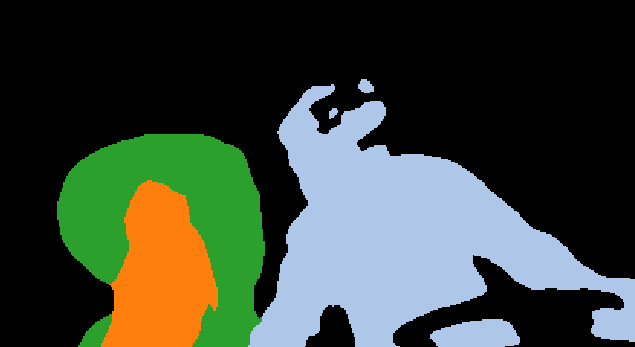}
 		\\

 		\rotatebox{90}{\tiny{DA}} \rotatebox{90}{\tiny{ShapeConv}} & \includegraphics[scale=0.1665]{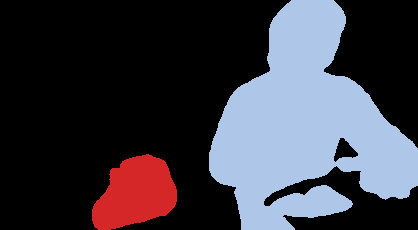}& 
 		\includegraphics[scale=0.1665]{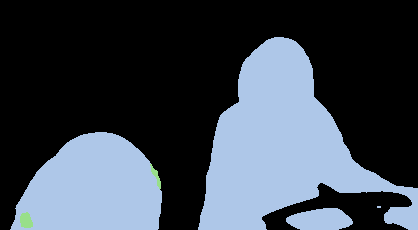}& 
 		\includegraphics[scale=0.1665]{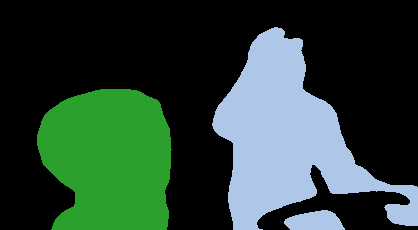}& 
 		\includegraphics[scale=0.1665]{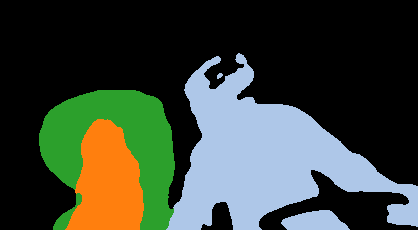}
 		\\
 		\rotatebox{90}{\tiny{MTL-DA}} \rotatebox{90}{\tiny{ShapeConv}} & \includegraphics[scale=0.07875]{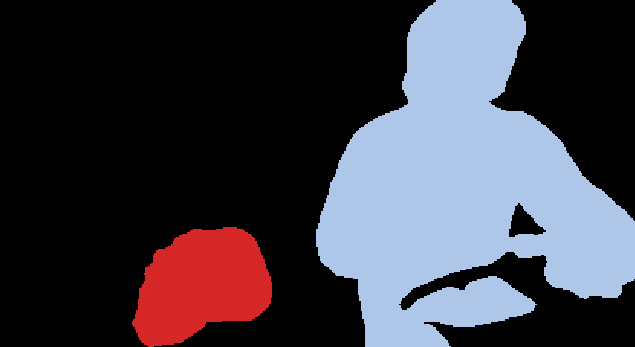}& 
 		\includegraphics[scale=0.07875]{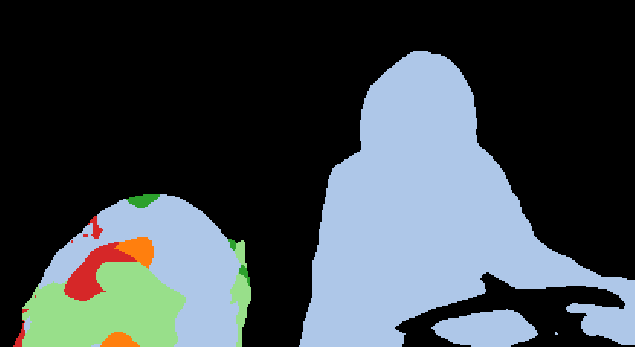}& 
 		\includegraphics[scale=0.07875]{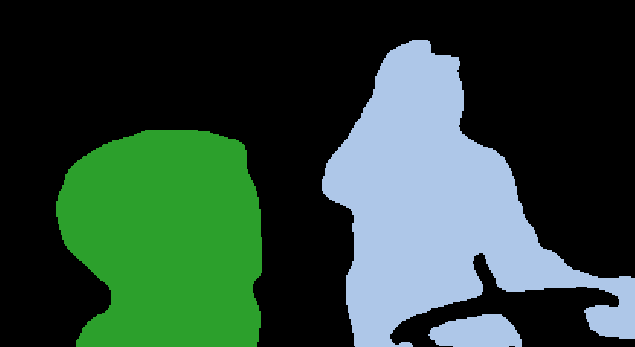}& 
 		\includegraphics[scale=0.07875]{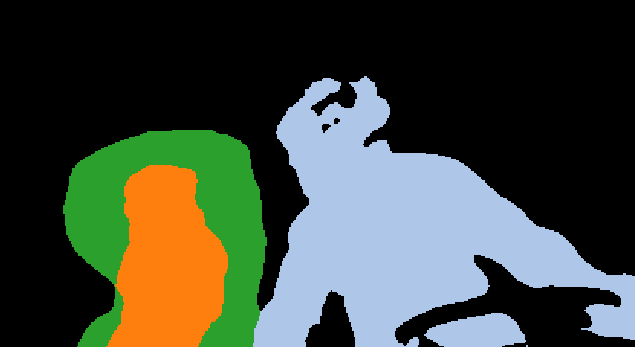}
 		\\

 	\end{tabular}
  \caption{Predictions from RGB-D and proposed I-D methods.}
  \label{predictionresults} 
\end{figure}

We can observe that the architectures that incorporate depth in an informed manner outperform FuseNet which simply concatenates depth with other modalities. Also, the combination of infrared and depth can be used instead of RGB-D input while achieving almost the same performance on segmentation but the disparity between the achieved performance is least when using depth-aware architectures with ShapeConv outperforming the other two baselines. We can see also from Figure \ref{predictionresults} that ShapeConv with both 3-channel and 1-channel infrared images have similar mask predictions.




\begin{table}[h]
\caption{Segmentation performance of suggested architectures. Input is always 2-channel concatenated IR-D image.}\label{traintest} \centering
\begin{tabular}{p{0.02cm}p{3.08cm}|p{0.65cm}|p{0.65cm}|p{0.65cm}|p{0.65cm}}
\hline
& \textbf{Method} & \textbf{Pixel Acc.} & \textbf{Class Acc.} & \textbf{Mean IoU} & \textbf{f.w.IoU} \\ 
\hline
& ShapeConv \cite{cao2021shapeconv} & \textbf{97.82} & 80.73 & 75.78 & \textbf{95.82} \\ 
\hline
\multirow{2}{*}{\rotatebox[origin=c]{90}{\textbf{Ours}}}
& DA-ShapeConv & 97.57 & 85.08 & 78.39 & 95.42 \\ 

& MTL-DA-ShapeConv & 97.73 & \textbf{85.98} & \textbf{79.73} & 95.68 \\ 
\hline
\end{tabular}
\label{novelresults}
\end{table}

\begin{table}[h]
\caption{Number of model parameters (in millions) and per image inference time (in milliseconds) for baselines and proposed MTL architecture. (Here, ch represents channels)}\label{traintest} \centering
\begin{tabular}{p{0.02cm}p{3.08cm}|p{0.9cm}|p{1.3cm}|p{1.0cm}}
\hline
& \textbf{Method} & \textbf{Input} & \textbf{Params (million)} & \textbf{Time (ms)} \\ 
\hline
& ShapeConv \cite{cao2021shapeconv} & 4-ch & 60.55 & 35.04 \\
& ShapeConv \cite{cao2021shapeconv} & 2-ch & 60.54 & 32.20 \\
\hline
\multirow{2}{*}{\rotatebox[origin=c]{90}{\textbf{Ours}}}
& DA-ShapeConv & 2-ch & 60.54  & 37.12 \\ 
& MTL-DA-ShapeConv & 2-ch & 78.12  & 37.16\\
\hline
\hline
\end{tabular}
\label{table-params}
\end{table}




\subsection{Experiments with Proposed Network}

We progressively modify the ShapeConv architecture into a multi-task learning network for semantic segmentation on IR-D data. We first replace the 3-channel infrared input with 1-channel infrared image, and evaluate the resulting network (Table \ref{novelresults} row 1). Comparing this with Table \ref{baselineresults} row four, we can notice slight improvement over most of the metrics while the number of model parameters decrease in the initial layers (Table \ref{table-params} row one and two). We follow this by incorporating depth similarity into the ShapeConv as described in section \ref{Depth-aware Shape Convolution} and name the realized method DA-ShapeConv (Depth Aware ShapeConv). We can note that combining both depth-aware and shape convolutions gives significant improvement on class accuracy as well as mean IoU. Finally, we train our proposed MTL network (MTL-DA-ShapeConv) and present the quantitative and qualitative results. We can see that the MTL architecture improve over the best performing RGB-D ShapeConv baseline from Table \ref{novelresults}, with significant improvement in class accuracy and mean IoU. This shows using our MTL network, one can outperform state-of-the-art RGB-D methods using images provided by a single ToF camera. 



\section{Conclusion}

We designed a network for IR-D segmentation that performs equally well as RGB-D segmentation so that inconvenient and expensive RGB-D cameras can be replaced with single Time-of-Flight (ToF) cameras. We showed that existing fusion approaches for RGB-D segmentation can be used with IR-D input if standard convolutions are replaced with depth-specific convolutions. We then presented a combination of depth-aware and shape-aware convolutions, and designed a multi-task learning (MTL) architecture with this new convolution operation. We employ hard parameter sharing between our main and auxiliary tasks of segmentation and depth filling respectively. Through progressive modifications to the input, the convolution operation, and the network architecture we showed that we can outperform all baseline methods. We conclude that using images from a single ToF camera, it is possible to surpass RGB-D segmentation performance with our designed MTL architecture.

\bibliographystyle{IEEEbib}

\end{document}